\documentclass{article}
\usepackage{iclr2027_conference,times}
\usepackage[T1]{fontenc}
\usepackage{amsmath,amssymb}
\usepackage{graphicx}
\usepackage{booktabs,array,longtable}
\usepackage{algorithm,algorithmic}
\usepackage{caption}
\usepackage{placeins}
\usepackage{xcolor}
\usepackage{colortbl}
\definecolor{basrowblue}{HTML}{D6EDF7}
\definecolor{basgroupgray}{HTML}{F0F0F0}
\newcolumntype{J}[1]{>{\raggedright\arraybackslash}m{#1}}
\newcolumntype{N}[1]{>{\centering\arraybackslash}m{#1}}
\usepackage[hyphens]{url}
\usepackage{hyperref}
\hypersetup{hidelinks,pdfauthor={Anonymous authors},
  pdftitle={BAS-OPD: Budget-Aware Selective On-Policy Self-Distillation for Fine-Grained Multimodal Perception}}
\graphicspath{{figures/}}
\newcolumntype{L}[1]{>{\raggedright\arraybackslash}p{#1}}
\newcolumntype{C}[1]{>{\centering\arraybackslash}p{#1}}
\newcolumntype{R}[1]{>{\raggedleft\arraybackslash}p{#1}}
\iclrfinalcopy
\hypersetup{pdfauthor={Zihan Chen, Hengguang Zhou, Yuan Kang, Yiming Zhang, Wenhui Fang, Zenghui Ding, Yining Sun, Cho-Jui Hsieh}}

\AddToHook{build/column/before}[bas-float-spacing]{%
  \ifnum\value{page}=\getpagerefnumber{fig:bas-framework}\relax
    \raggedbottom
  \fi
  \ifnum\value{page}=\getpagerefnumber{tab:training-comparison}\relax
    \raggedbottom
  \fi
  \ifnum\value{page}=\getpagerefnumber{fig:utility-quantile}\relax
    \raggedbottom
  \fi
}
\AddToHook{build/column/after}[bas-float-spacing]{%
  \ifnum\value{page}=\getpagerefnumber{fig:bas-framework}\relax
    \flushbottom
  \fi
  \ifnum\value{page}=\getpagerefnumber{tab:training-comparison}\relax
    \flushbottom
  \fi
  \ifnum\value{page}=\getpagerefnumber{fig:utility-quantile}\relax
    \flushbottom
  \fi
}
\makeatletter
\AddToHook{cmd/appendix/after}[bas-appendix-floats]{%
  \setlength{\@fptop}{0pt}%
  \setlength{\@fpsep}{14pt}%
  \setlength{\@fpbot}{0pt plus 1fil}%
}
\makeatother

\title{\raggedright BAS-OPD: Budget-Aware Selective\\
On-Policy Self-Distillation for\\
Fine-Grained Multimodal Perception}
\author{\hspace*{-\tabcolsep}\begin{minipage}{\dimexpr\textwidth-\tabcolsep\relax}\raggedright
Zihan Chen$^{1,2}$,\enspace Hengguang Zhou$^{3}$,\enspace
Yuan Kang$^{1,2}$,\enspace Yiming Zhang$^{1,2}$,\\[2pt]
Wenhui Fang$^{1,2}$,\enspace Zenghui Ding$^{1}$,\enspace
Yining Sun$^{1,2}$,\enspace Cho-Jui Hsieh$^{3}$\\[4pt]
{\normalfont\small
$^{1}$HFIPS, Chinese Academy of Sciences\hspace{0.8em}%
$^{2}$University of Science and Technology of China\\
$^{3}$University of California, Los Angeles}
\end{minipage}}

\begin{document}
\maketitle
\lhead{Preprint}
\par\noindent
\begin{minipage}{\textwidth}
    \centering
    \phantomsection\label{fig:table1-overview}
    \includegraphics[width=\textwidth]{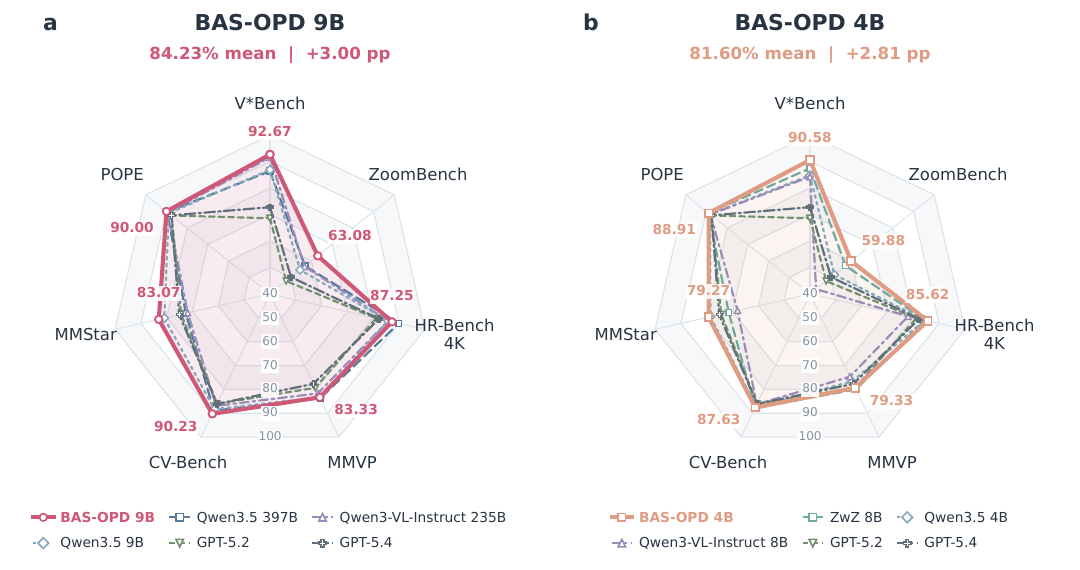}
    \captionsetup{font=small,skip=3pt}
    \captionof*{figure}{Accuracy (\%) across seven visual understanding benchmarks: V*Bench, ZoomBench, HR-Bench 4K, MMVP, CV-Bench, MMStar and POPE.}
\end{minipage}
\par\medskip

\begin{abstract}
Multimodal large language models (MLLMs) often struggle with fine-grained visual perception when processing complete images, as critical evidence may only appear in local regions. On-policy self-distillation (OPD) enables transferring privileged visual knowledge from informative views to full-image policies, but querying the teacher for every rollout introduces substantial supervision costs. In this work, we propose BAS-OPD, a budget-aware selective OPD framework that allocates teacher supervision under limited query budgets. Instead of querying all rollouts, BAS-OPD selects informative samples while maintaining full-batch student generation. We explore random, uncertainty-based, and learned utility-based selection strategies, where the learned selector estimates query value from detached rollout statistics and online utility signals derived from student--teacher agreement and teacher confidence without additional student forward passes. BAS-OPD only changes training-time supervision allocation and preserves single-pass full-image inference. Experiments on fine-grained multimodal perception benchmarks demonstrate that BAS-OPD achieves strong performance while substantially reducing teacher supervision costs, highlighting the effectiveness of selective OPD under constrained budgets.
\end{abstract}

\section{Introduction}

Multimodal large language models (MLLMs) have become general-purpose interfaces for combining visual perception with language reasoning. Their performance, however, remains brittle when an answer depends on a small object, a local attribute, or a spatial relation embedded in a high-resolution scene. V*Bench and HR-Bench expose this limitation by requiring models to locate and interpret fine visual evidence in large images \citep{wu2024vstar,wang2025dc2}. The underlying difficulty is not only local recognition. Resizing an image or encoding the scene at nearly uniform spatial fidelity can weaken the few visual signals that determine the answer. Reliable fine-grained perception must therefore preserve local evidence while directing limited computation toward the regions that matter.

Existing methods improve access to fine visual evidence along two main directions. Model-side approaches retain more pixels through high-resolution encoders or dynamic tiling; for example, InternVL~1.5 processes an image as up to 40 tiles to support 4K inputs \citep{chen2024internvl}. Inference-time approaches instead decide where to look. V* performs language-guided visual search, while DC$^2$ recursively partitions an image, describes its patches, and retrieves relevant regional information \citep{wu2024vstar,wang2025dc2}. These methods reduce information loss, but they trade against larger visual-token budgets, repeated patch processing, or additional inference calls. More importantly, they recover detail as an inference-time operation rather than teaching a single-pass full-image policy to internalize the advantage of focused regional evidence.

\begin{figure}[t]
    \centering
    \includegraphics[width=\textwidth]{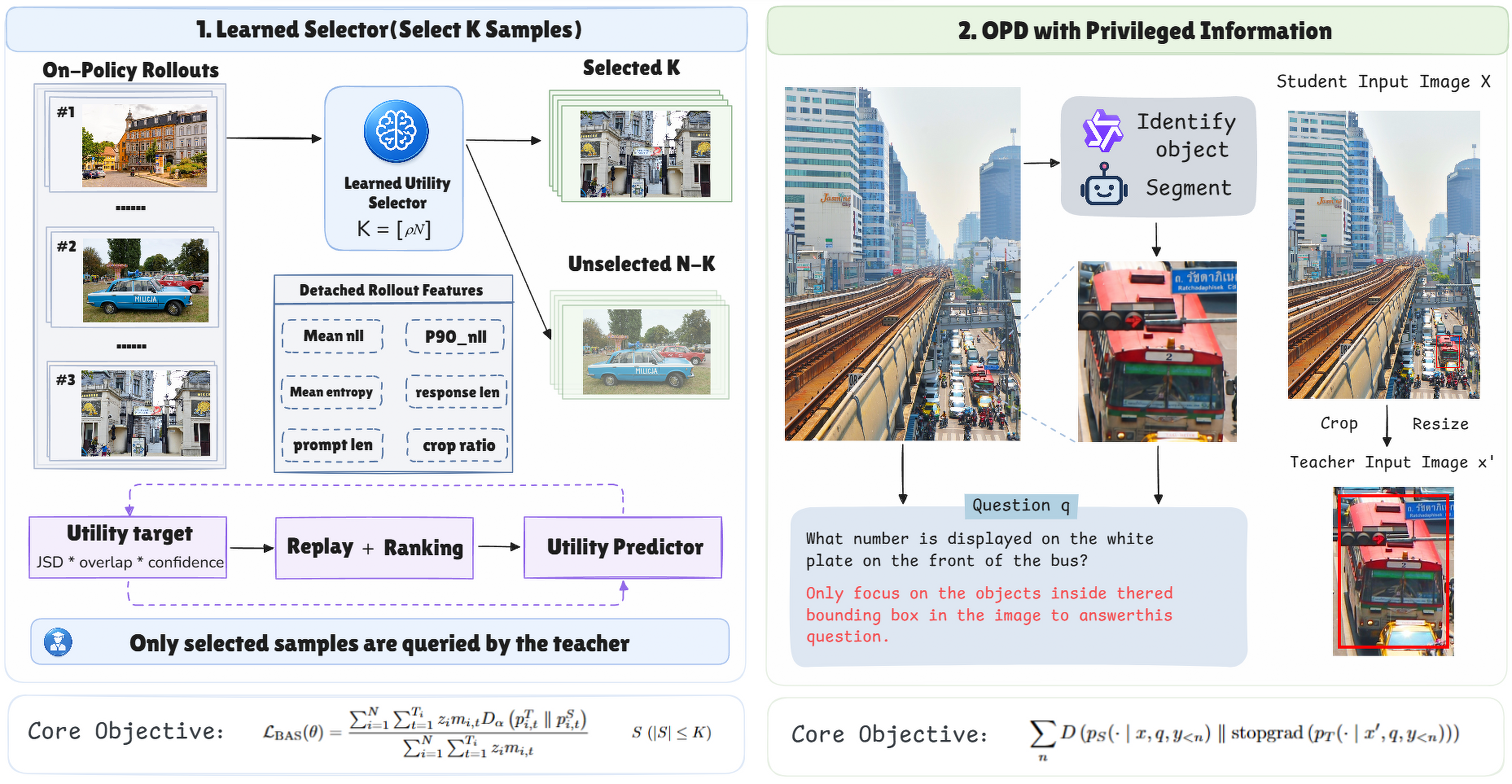}
    \caption{Overview of BAS-OPD. The full-image policy generates all rollouts; a budget-aware selector sends only $K$ samples to the crop teacher for distillation and utility learning. Selection is removed at inference.}
    \label{fig:bas-framework}
\end{figure}

On-policy distillation transfers teacher behavior along trajectories sampled from the student. In autoregressive language models, it reduces the mismatch between teacher-forced training and student-generated inference while supporting sequence-level knowledge transfer \citep{agarwal2024gkd,ko2024distillm}. A recent multimodal formulation applies this principle to transfer privileged regional evidence into a single-pass full-image policy \citep{yuan2026visionopd}. Existing methods mainly focus on learning objectives or supervision quality and typically query the teacher for every eligible rollout. This ties supervision cost to the full rollout batch, even when feedback is redundant or weakly informative. The unresolved challenge is to allocate teacher queries selectively under a strict budget while preserving the benefits of student-generated training.

We introduce BAS-OPD, which selects on-policy rollouts before crop-teacher scoring under a strict sample-level budget. The full-image policy generates responses for the entire batch, but only selected samples enter teacher scoring and distillation. We support three selection strategies: random, student-entropy, and learned utility selection. The learned selector uses detached rollout statistics, requiring no additional student forward pass. Online utility targets combine student--teacher divergence, top-$k$ overlap, and teacher confidence to prioritize large, confident corrections from privileged regional evidence. Selection is used only during training, preserving single-pass full-image inference.

We evaluate BAS-OPD with Qwen3.5-4B and Qwen3.5-9B on seven benchmarks, measuring downstream accuracy together with teacher calls and scored tokens. With the 4B backbone, learned selection under a 25\% teacher-call budget achieves an average accuracy of 81.60\%, 1.84 points above full querying, while using only 15.78\% of its teacher-scored tokens. Scaling BAS-OPD to 9B yields an average accuracy of 84.23\%, 3.00 points above the strongest non-BAS open-source single-forward-pass model. Within this comparison group, BAS-OPD 9B ranks first on five benchmarks and second on the remaining two. A frozen-selector audit further shows that learned selection retrieves higher-utility rollouts than student-entropy selection. Our contributions are:

\noindent\textbullet\enspace \textbf{Budget-Aware OPD.} We formulate regional-to-global on-policy self-distillation with a strict, directly measurable teacher-query budget.

\noindent\textbullet\enspace \textbf{Online Utility Selection.} We unify random, student-entropy, and learned utility policies while ensuring that unselected rollouts never enter the crop-teacher forward pass.

\noindent\textbullet\enspace \textbf{Accuracy--Cost and Selection Diagnostics.} We evaluate accuracy and teacher cost across policies and budgets, with utility, ranking, and gradient analyses characterizing how learned selection concentrates supervision.

\section{Method}

\newcommand{\selectorcomponentstable}{%
\begin{table}[!htbp]
\centering
\caption{Component ablations of the learned utility selector across seven benchmarks. Avg. is the unweighted mean across the seven benchmarks.}
\label{tab:selector-components}
\fontsize{9.5}{10.8}\selectfont
\renewcommand{\arraystretch}{1.12}
\setlength{\tabcolsep}{1.3pt}
\setlength{\aboverulesep}{1.2pt}
\setlength{\belowrulesep}{1.2pt}
\begin{tabular}{
J{\dimexpr0.318\textwidth-2\tabcolsep\relax}
N{\dimexpr0.095\textwidth-2\tabcolsep\relax}
N{\dimexpr0.067\textwidth-2\tabcolsep\relax}
N{\dimexpr0.109\textwidth-2\tabcolsep\relax}
N{\dimexpr0.080\textwidth-2\tabcolsep\relax}
N{\dimexpr0.106\textwidth-2\tabcolsep\relax}
N{\dimexpr0.092\textwidth-2\tabcolsep\relax}
N{\dimexpr0.071\textwidth-2\tabcolsep\relax}
N{\dimexpr0.062\textwidth-2\tabcolsep\relax}
}
\toprule
& \multicolumn{3}{c}{\small\textbf{Fine-Grained Visual Tasks}} & \multicolumn{4}{c}{\small\textbf{Holdout Tasks}} & \\
\cmidrule(lr){2-4}\cmidrule(lr){5-8}
\small\textbf{Selector variant} & \small\textbf{V*Bench} & \small\textbf{\shortstack{Zoom\\Bench}} & \small\textbf{\shortstack{HR-Bench\\4K}} & \small\textbf{MMVP} & \small\textbf{CV-Bench} & \small\textbf{MMStar} & \small\textbf{POPE} & \small\textbf{Avg.} \\
\midrule
Without top-$k$ overlap ($d\,c$) & 86.39 & 57.04 & 82.25 & 76.33 & 86.09 & 70.80 & 88.97 & 78.27 \\
Without teacher confidence ($d\,o$) & 88.48 & 55.50 & 81.00 & 77.33 & 84.99 & 69.40 & 88.87 & 77.94 \\
Without ranking loss & 89.53 & 56.69 & 79.38 & 76.33 & 85.46 & 70.53 & \textbf{89.11} & 78.15 \\
Without random exploration & 85.86 & 58.70 & 83.38 & 77.00 & 86.23 & 71.47 & 89.03 & 78.81 \\
\midrule
\rowcolor{basrowblue}
\textbf{Full learned selector ($d\,o\,c$)} & \textbf{90.58} & \textbf{59.88} & \textbf{85.62} & \textbf{79.33} & \textbf{87.63} & \textbf{79.27} & 88.91 & \textbf{81.60} \\
\bottomrule
\end{tabular}

\end{table}
}

\subsection{Budget-Aware Selective OPD}

Figure~\ref{fig:bas-framework} summarizes the training and inference paths. Algorithm~\ref{alg:bas-opd} details the learned-selection training loop. BAS-OPD augments regional-to-global OPD with an explicit budget on crop-teacher supervision \citep{yuan2026visionopd}. Consider a training batch $\mathcal{B}=\{(x_i,\tilde{x}_i,q_i)\}_{i=1}^{N}$, where $x_i$ is the full image, $\tilde{x}_i$ is its privileged evidence-centered crop, and $q_i$ is the language prompt. The full-image policy first samples an on-policy response $y_i=(y_{i,1},\ldots,y_{i,T_i})$ for every example in the batch. For response position $t$, we denote the student distribution conditioned on the full image by $p^S_{i,t}$ and the crop-teacher distribution over the same response prefix by $p^T_{i,t}$:
\begin{align}
p^{S}_{i,t}&=p_{\theta}(\cdot\mid x_i,q_i,y_{i,<t}),\\
p^{T}_{i,t}&=p_{\bar{\theta}}(\cdot\mid \tilde{x}_i,q_i,y_{i,<t}).
\end{align}
Here, $\theta$ denotes the trainable full-image policy and $\bar{\theta}$ denotes the crop-conditioned teacher maintained by the underlying OPD procedure. Standard OPD minimizes the expected token-level divergence along responses sampled from the student:
\begin{equation}
\mathcal{L}_{\mathrm{OPD}}(\theta)
=\mathbb{E}_{\substack{(x_i,\tilde{x}_i,q_i)\sim\mathcal{D}\\
y_i\sim p_{\theta}(\cdot\mid x_i,q_i)}}
\left[
\frac{1}{T_i}\sum_{t=1}^{T_i}
D_{\alpha}\!\left(p^{T}_{i,t}\,\Vert\,p^{S}_{i,t}\right)
\right],
\label{eq:opd-objective}
\end{equation}
where $\mathcal{D}$ is the training distribution and $D_{\alpha}$ is the distillation divergence; our experiments use Jensen--Shannon divergence with $\alpha=0.5$. The teacher scores the same response prefixes generated by the student. BAS-OPD does not change how responses are generated or how the teacher is regularized; it changes which rollouts are allowed to enter the crop-teacher forward pass.

Let $\mathcal{C}\subseteq\{1,\ldots,N\}$ be the samples for which privileged crops are available. Given a query ratio $\rho\in(0,1]$, we impose the sample budget
\begin{equation}
K=\left\lfloor\rho\lvert\mathcal{C}\rvert\right\rfloor,
\qquad
\mathcal{S}\subseteq\mathcal{C},\quad \lvert\mathcal{S}\rvert\leq K,
\end{equation}
where $\mathcal{S}$ is chosen before any crop-teacher computation. The full-image policy therefore produces rollouts for the complete batch, while the crop teacher scores only the selected subset. This ordering is important: the budget reduces the actual teacher input batch rather than masking losses after an already completed teacher forward pass. For $i\notin\mathcal{S}$, no crop-teacher distribution or OPD target is constructed.

Selection aims to maximize the total utility of the queried subset under this budget. Because utility is observed only after teacher scoring, we predict it from statistics already produced by student rollouts (Algorithm~\ref{alg:bas-opd}, lines~\ref{alg:rollout}--\ref{alg:select-subset}).

\begin{algorithm}[!t]
\caption{BAS-OPD with learned utility selection}
\label{alg:bas-opd}
\small
\renewcommand{\algorithmicrequire}{\textbf{Input:}}
\renewcommand{\algorithmicensure}{\textbf{Output:}}
\algsetup{indent=1.1em,linenosize=\scriptsize}
\begin{algorithmic}[1]
\REQUIRE Policy $\theta$; query ratio $\rho$; exploration ratio $\epsilon$; replay threshold $M$; EMA rate $\tau$.
\ENSURE Trained full-image policy $\theta$ for inference without crops or a selector.
\STATE Initialize $\bar\theta\leftarrow\theta$, predictor $f_\psi$, feature normalizer, and FIFO buffer $\mathcal{R}\leftarrow\varnothing$.
\FOR{each training step $s$ with rollout batch $\mathcal{B}$}
    \STATE Generate $y_i\sim p_\theta(\cdot\mid x_i,q_i)$ for all $i\in\mathcal{B}$; identify crop-eligible candidates $\mathcal{C}$.\label{alg:rollout}
    \STATE Form detached features $\phi_i$ (Eq.~\ref{eq:bas-features}) and update normalization on $\mathcal{C}$; set $K\leftarrow\lfloor\rho|\mathcal{C}|\rfloor$.\label{alg:features}
    \IF{$f_\psi$ has not yet been updated}\label{alg:selection}
        \STATE $\mathcal{S}\leftarrow\operatorname{Uniform}(\mathcal{C},K)$, sampling without replacement.
    \ELSE
        \STATE Predict $\hat u_i\leftarrow f_\psi(\operatorname{Normalize}(\phi_i))$ for $i\in\mathcal{C}$; set $K_{\mathrm e}\leftarrow\lceil\epsilon K\rceil$.
        \STATE Form $\mathcal{S}$ from the top $K-K_{\mathrm e}$ candidates by $\hat u_i$ plus $K_{\mathrm e}$ candidates sampled uniformly without replacement from the remainder.\label{alg:select-subset}
    \ENDIF
    \STATE Run the standard OPD student forward for $p^S$; obtain detached crop-teacher $p^T$ on the same tokens \textbf{only for} $i\in\mathcal{S}$.\label{alg:teacher}
    \STATE Compute detached utility labels $u_i$ for $i\in\mathcal{S}$ (Eq.~\ref{eq:bas-utility}).\label{alg:utility}
    \STATE Update $\theta$ using $\mathcal{L}_{\mathrm{BAS}}$ (Eq.~\ref{eq:bas-distillation}); after a valid update, set $\bar\theta\leftarrow(1-\tau)\bar\theta+\tau\theta$.\label{alg:distill}
    \STATE Add valid detached $(\phi_i,u_i,s)$ to bounded replay $\mathcal{R}$; if new pairs were added and $|\mathcal{R}|\ge M$, update \textbf{only} $\psi$ from replay using $\mathcal{L}_{\mathrm{sel}}$ (Eq.~\ref{eq:bas-selector-loss}).\label{alg:selector-update}
\ENDFOR
\end{algorithmic}
\end{algorithm}

\subsection{Learned Utility Selection}

For each rollout, Algorithm~\ref{alg:bas-opd} (line~\ref{alg:features}) constructs a detached feature vector
\begin{equation}
\phi_i=
\left[
\overline{\ell}_i,
Q_{0.9}(\ell_i),
\overline{H}_i,
T_i,
L_i,
a_i
\right]^{\top},
\label{eq:bas-features}
\end{equation}
which summarizes rollout likelihood and entropy, response and prompt lengths, and crop area. These quantities require no additional MLLM forward pass. A lightweight predictor estimates query utility from normalized features; feature definitions, running normalization, and network architecture are given in Appendix~\ref{app:bas-features}:
\begin{equation}
\hat{u}_i=f_{\psi}(\operatorname{Normalize}(\phi_i)).
\end{equation}

Algorithm~\ref{alg:bas-opd} (lines~\ref{alg:selection}--\ref{alg:select-subset}) selects $K$ candidates uniformly before the first predictor update. Thereafter, it reserves $K_{\mathrm{e}}=\lceil\epsilon K\rceil$ queries for random exploration and assigns the remaining $K-K_{\mathrm{e}}$ queries to the highest-scoring candidates. Exploration samples from the remaining candidates to avoid a self-confirming selection loop while preserving the strict budget.

The supervision for $f_{\psi}$ is obtained online from selected rollouts (Algorithm~\ref{alg:bas-opd}, line~\ref{alg:utility}). Let $P^{S}_{i,t}$ and $P^{T}_{i,t}$ denote student and teacher probabilities evaluated on the student's top-$k$ token indices and renormalized over this shared support. We measure their disagreement with the Jensen--Shannon divergence
\begin{equation}
d_{i,t}=\operatorname{JSD}\!\left(P^{S}_{i,t},P^{T}_{i,t}\right).
\end{equation}
Divergence alone can be uninformative when the teacher is diffuse or favors different tokens. We define $o_{i,t}$ as the fraction of student top-$k$ indices also in the teacher's own top-$k$ set, and compute the normalized teacher confidence
\begin{equation}
c_{i,t}=1-\frac{H(P^{T}_{i,t})}{\log k},
\end{equation}
and define the detached per-sample target as
\begin{equation}
u_i=
\frac{1}{\sum_t m_{i,t}}
\sum_t m_{i,t}\,d_{i,t}\,o_{i,t}\,c_{i,t},
\label{eq:bas-utility}
\end{equation}
where $m_{i,t}$ masks padding and invalid response positions. This target is high when the crop teacher makes a large, support-consistent, and confident correction to the full-image policy. Only queried samples yield $(\phi_i,u_i)$ pairs; unselected samples are never assigned synthetic utility labels.

Observed feature--utility pairs train the predictor through replay using weighted Huber regression and pairwise ranking:
\begin{equation}
\mathcal{L}_{\mathrm{sel}}
=\mathcal{L}_{\mathrm{reg}}+\lambda_{\mathrm{rank}}\mathcal{L}_{\mathrm{rank}},
\label{eq:bas-selector-loss}
\end{equation}
Regression calibrates utility values, while ranking learns their ordering. Algorithm~\ref{alg:bas-opd} (line~\ref{alg:selector-update}) updates only the selector from detached features and targets; the replay procedure and loss definitions are given in Appendix~\ref{app:selector-training}.

\subsection{Selective Distillation and Inference}

Algorithm~\ref{alg:bas-opd} (lines~\ref{alg:teacher}--\ref{alg:distill}) scores only $\mathcal{S}$ with the crop teacher and compares its detached distributions with the standard OPD student distributions on the same response tokens. Using $D_{\alpha}$ from Eq.~\ref{eq:opd-objective}, we aggregate selected valid tokens with a masked mean. With $z_i=\mathbf{1}[i\in\mathcal{S}]$, the budgeted objective is
\begin{equation}
\mathcal{L}_{\mathrm{BAS}}(\theta)
=\frac{
\sum_{i=1}^{N}\sum_{t=1}^{T_i}
z_i m_{i,t}\,
D_{\alpha}\!\left(p^{T}_{i,t}\,\Vert\,p^{S}_{i,t}\right)
}{
\max\!\left(1,\sum_{i=1}^{N}\sum_{t=1}^{T_i}z_i m_{i,t}\right)
}.
\label{eq:bas-distillation}
\end{equation}
At full budget, $\rho=1$ and $\mathcal{S}=\mathcal{C}$, this objective recovers full-query OPD with the same token-mean reduction. Under a reduced budget, unselected rows are absent from the crop-teacher forward pass and are masked out of the distillation loss. Empty-selection and update handling are detailed in Appendix~\ref{app:selector-training}. The MLLM parameters $\theta$ and selector parameters $\psi$ are optimized separately by $\mathcal{L}_{\mathrm{BAS}}$ and $\mathcal{L}_{\mathrm{sel}}$, respectively. Baselines and supervision-cost definitions are given in Appendices~\ref{app:selection-baselines} and~\ref{app:teacher-cost}. All selection components are training-only. At inference, BAS-OPD retains the original full-image policy and produces an answer in a single forward pass without privileged crops, a selector, or an external visual tool.

\section{Experiments}

\subsection{Experimental Setup}

\noindent\textbf{MLLM and selector training.}
We evaluate BAS-OPD with Qwen3.5-4B and Qwen3.5-9B; policy ablations and cost measurements use the 4B model. Training uses 3,120 preprocessed full-image--crop pairs from the released training set \citep{yuan2026visionopd}, JSD with $\alpha=0.5$, top-$100$ distillation, a crop-teacher EMA rate of 0.05, and maximum prompt/response lengths of 8,192/1,024 tokens. The MLLM parameters $\theta$ use AdamW (lr $2\times10^{-6}$; weight decay 0.01); the selector parameters $\psi$ use separate Adam (lr $1\times10^{-3}$) in Learned-10/25/50. Each run uses 28 prompts per step and eight rollouts per prompt, trains for one epoch on four NVIDIA A100-SXM4 80\,GB GPUs, and uses training seed 42. Appendix~\ref{app:bas-features} and Table~\ref{tab:supp-training-config} give the remaining selector settings and run lengths.

\noindent\textbf{Evaluation benchmarks.}
We evaluate on seven benchmarks spanning high-resolution perception, general visual understanding, and object hallucination. V*Bench \citep{wu2024vstar}, ZoomBench \citep{wei2026zooming}, and HR-Bench 4K \citep{wang2025dc2} assess fine-grained perception; MMVP \citep{tong2024eyes}, CV-Bench \citep{tong2024cambrian}, and MMStar \citep{chen2024mmstar} measure broader visual capabilities. POPE \citep{li2023pope} evaluates object hallucination across adversarial, popular, and random splits.

\begin{table}[!htbp]
\centering
\caption{Comparison with state-of-the-art MLLMs across seven benchmarks. We report accuracy (\%) for each benchmark. Among the reported open-source single-forward-pass results, the best value is shown in bold and the second-best is underlined.}
\label{tab:sota-comparison}
\fontsize{9.5}{10.8}\selectfont
\renewcommand{\arraystretch}{1.06}
\setlength{\tabcolsep}{1.3pt}
\setlength{\aboverulesep}{1.2pt}
\setlength{\belowrulesep}{1.2pt}
\begin{tabular}{
J{\dimexpr0.198\textwidth-2\tabcolsep\relax}
N{\dimexpr0.085\textwidth-2\tabcolsep\relax}
N{\dimexpr0.100\textwidth-2\tabcolsep\relax}
N{\dimexpr0.135\textwidth-2\tabcolsep\relax}
N{\dimexpr0.115\textwidth-2\tabcolsep\relax}
N{\dimexpr0.087\textwidth-2\tabcolsep\relax}
N{\dimexpr0.113\textwidth-2\tabcolsep\relax}
N{\dimexpr0.096\textwidth-2\tabcolsep\relax}
N{\dimexpr0.071\textwidth-2\tabcolsep\relax}
}
\toprule
\small\textbf{Model} & \small\textbf{\shortstack{Param.\\Size}} & \small\textbf{V*Bench} & \small\textbf{ZoomBench} & \small\textbf{\shortstack{HR-Bench\\4K}} & \small\textbf{MMVP} & \small\textbf{CV-Bench} & \small\textbf{MMStar} & \small\textbf{POPE} \\
\midrule
\multicolumn{9}{c}{\textbf{\textit{``Thinking-with-Images'' Agentic Models}}} \\
\midrule
DeepEyes & 7B & 79.58 & 46.04 & 74.25 & 72.00 & 76.81 & 61.33 & 88.42 \\
Thyme & 7B & 81.68 & 46.39 & 76.75 & 70.33 & 75.81 & 60.53 & 60.53 \\
DeepEyesV2 & 7B & 80.63 & 46.27 & 75.00 & 72.33 & 80.48  & 61.07 & 89.08 \\
\midrule

\multicolumn{9}{c}{\textbf{\textit{Closed-Source Models (Single Forward Pass)}}} \\
\midrule
GPT-5.2 & -- & 68.59 & 47.81 & 81.12 & 79.33 & 85.99 & 74.40 & 87.51 \\
GPT-5.4 & -- & 72.77 & 50.18 & 82.25 & 77.67 & 86.08 & 75.13 & 87.58 \\
Gemini-3.1-Pro & -- & 88.48 & 60.83 & 89.62 & 91.00 & 89.98 & 83.67 & 89.72 \\
Gemini-3.5-Flash & -- & 90.05 & 60.95 & 89.50 & 94.33 & 88.83 & 84.20 & 89.88 \\
\midrule

\multicolumn{9}{c}{\textbf{\textit{Open-Source Models (Single Forward Pass)}}} \\
\midrule
Qwen3.5 & 4B & 84.82 & 51.72 & 84.38 & 76.67 & 87.13 & 78.53 & 88.28 \\
Qwen3.5 & 9B & 86.91 & 54.56 & 85.75 & \underline{83.33} & 88.29 & \underline{80.93} & 88.88 \\
Qwen3-VL-Instruct & 8B & 84.29 & 42.96 & 78.00 & 74.67 & 86.22 & 68.07 & 88.12 \\
MiMo-VL-RL & 7B & 81.68 & 43.20 & 73.25 & 75.67 & 80.45 & 69.87 & 88.36 \\
ZwZ & 8B & 87.43 & 57.16 & 84.50 & 80.00 & 87.77 & 71.47 & 88.97 \\
MiniCPM-V-4.5 & 9B & 69.63 & 42.84 & 68.75 & 78.00 & 79.88 & 67.53 & 87.51 \\
Qwen3-VL-Instruct & 235B & \underline{91.62} & 56.69 & 85.75 & 81.67 & 86.90 & 71.87 & 89.26 \\
Qwen3.5 & 397B & 86.39 & 56.92 & \textbf{89.88} & \textbf{83.67} & \underline{88.92} & 72.60 & \underline{89.63} \\
\midrule

\rowcolor{basrowblue}
\textbf{BAS-OPD (4B)} & 4B & 90.58 & \underline{59.88} & 85.62 & 79.33 & 87.63 & 79.27 & 88.91 \\
\rowcolor{basrowblue}
\textbf{BAS-OPD (9B)} & 9B & \textbf{92.67} & \textbf{63.08} & \underline{87.25} & \underline{83.33} & \textbf{90.23} & \textbf{83.07} & \textbf{90.00} \\
\bottomrule

\end{tabular}

\end{table}

\noindent\textbf{Evaluation protocol.}
Evaluation uses only the original full image. Privileged crops and all selection modules are disabled. We use deterministic decoding with temperature 0 and seed 42. Benchmark-specific parsers score multiple-choice and exact-answer outputs. GPT-5.5 at temperature 0 judges the remaining free-form responses. We report accuracy or the official aggregate score for each benchmark.

\subsection{Experimental Results}

\noindent\textbf{Comparison with state-of-the-art MLLMs.}
Table~\ref{tab:sota-comparison} compares BAS-OPD with agentic, closed-source, and open-source MLLMs. BAS-OPD 9B averaged 84.23\%, exceeding Qwen3.5-9B, the strongest non-BAS open-source single-forward-pass model by average score, by 3.00 points. Among open-source single-forward-pass models, it ranked first on five benchmarks and second on two. It also outperformed Qwen3-VL-Instruct 235B on all seven benchmarks. BAS-OPD 4B averaged 81.60\%, exceeding Qwen3.5-9B and Qwen3.5-397B by 0.37 and 0.46 points, respectively. Against closed-source models, BAS-OPD 9B led on four benchmarks, while Gemini remained stronger on the other three.

\FloatBarrier

\noindent\textbf{Comparison with alternative training objectives.}
Table~\ref{tab:training-comparison} compares BAS-OPD with vanilla inference, SFT, GRPO, DAPO, and OPSD on matched Qwen3.5 backbones. At 4B and 9B, BAS-OPD led on six benchmarks, tied the best MMVP score, and improved the OPSD average by 3.03 and 3.24 points, respectively, under a 25\% teacher-query budget. Its largest gains over the strongest alternatives were on V*Bench and ZoomBench.

\begingroup
\setlength{\intextsep}{9pt}
\begin{table}[H]
\centering
\caption{Accuracy (\%) of vanilla Qwen3.5 \citep{qwen2026qwen35}, SFT \citep{ouyang2022training}, GRPO \citep{shao2024deepseekmath}, DAPO \citep{yu2025dapo}, OPSD \citep{zhao2026selfdistilled}, and BAS-OPD (learned selection, 25\% teacher-query budget). Best results per backbone are bold.}
\label{tab:training-comparison}
\fontsize{9.5}{10.8}\selectfont
\renewcommand{\arraystretch}{1.06}
\setlength{\tabcolsep}{1.3pt}
\setlength{\aboverulesep}{1.2pt}
\setlength{\belowrulesep}{1.2pt}
\begin{tabular}{
J{\dimexpr0.250\textwidth-2\tabcolsep\relax}
N{\dimexpr0.100\textwidth-2\tabcolsep\relax}
N{\dimexpr0.135\textwidth-2\tabcolsep\relax}
N{\dimexpr0.120\textwidth-2\tabcolsep\relax}
N{\dimexpr0.100\textwidth-2\tabcolsep\relax}
N{\dimexpr0.115\textwidth-2\tabcolsep\relax}
N{\dimexpr0.100\textwidth-2\tabcolsep\relax}
N{\dimexpr0.080\textwidth-2\tabcolsep\relax}
}
\toprule
& \multicolumn{3}{c}{\textbf{Fine-Grained Visual Tasks}} & \multicolumn{4}{c}{\textbf{Holdout Tasks}} \\
\cmidrule(lr){2-4}\cmidrule(lr){5-8}
\textbf{Method} & \textbf{V*Bench} & \textbf{ZoomBench} & \textbf{\shortstack{HR-Bench\\4K}} & \textbf{MMVP} & \textbf{CV-Bench} & \textbf{MMStar} & \textbf{POPE} \\
\midrule
\rowcolor{basgroupgray}
\multicolumn{8}{c}{\textit{Qwen3.5-4B}} \\
Vanilla & 84.82 & 51.72 & 84.38 & 76.67 & 87.13 & 78.53 & 88.28 \\
\midrule
SFT on Self-Teacher & 78.53 & 54.44 & 80.00 & 78.00 & 85.78 & 68.27 & 87.44 \\
GRPO & 83.25 & 55.62 & 82.50 & \textbf{79.33} & 87.16 & 70.67 & 86.39 \\
DAPO & 85.86 & 55.38 & 84.38 & \textbf{79.33} & 87.07 & 72.53 & 86.59 \\
OPSD & 83.77 & 54.32 & 82.00 & 79.00 & 87.19 & 74.87 & 88.88 \\
\rowcolor{basrowblue}
\textbf{BAS-OPD} & \textbf{90.58} & \textbf{59.88} & \textbf{85.62} & \textbf{79.33} & \textbf{87.63} & \textbf{79.27} & \textbf{88.91} \\
\midrule
\rowcolor{basgroupgray}
\multicolumn{8}{c}{\textit{Qwen3.5-9B}} \\
Vanilla & 86.91 & 54.56 & 85.75 & \textbf{83.33} & 88.29 & 80.93 & 88.88 \\
\midrule
SFT on Self-Teacher & 81.68 & 58.82 & 83.13 & 81.33 & 88.05 & 73.53 & 87.50 \\
GRPO & 85.86 & 57.40 & 87.00 & 80.67 & 87.70 & 73.33 & 87.72 \\
DAPO & 87.43 & 56.09 & 86.00 & 79.67 & 87.39 & 75.80 & 87.51 \\
OPSD & 90.58 & 57.75 & 84.50 & 80.33 & 87.37 & 78.93 & 87.50 \\
\rowcolor{basrowblue}
\textbf{BAS-OPD} & \textbf{92.67} & \textbf{63.08} & \textbf{87.25} & \textbf{83.33} & \textbf{90.23} & \textbf{83.07} & \textbf{90.00} \\
\bottomrule
\end{tabular}

\end{table}
\endgroup

\FloatBarrier
\subsection{Ablation and Efficiency Analysis}

\noindent\textbf{Accuracy--cost trade-off.}
Table~\ref{tab:budget-ablation} reports seven-benchmark accuracy and teacher cost. Figure~\ref{fig:accuracy-cost} in the appendix visualizes the comparison for full querying, Entropy-25, and Learned-25. Full querying averaged 79.76\%, versus 79.25\%, 80.24\%, and 81.60\% for Random-25, Entropy-25, and Learned-25. Learned-25 improved accuracy by 2.35, 1.36, and 1.84 points over random selection, entropy selection, and full querying, respectively. The three budgeted policies used identical query ratios, isolating how teacher calls were allocated. Learned-25 exceeded full querying on every benchmark, with the largest gains on HR-Bench 4K (3.37 points) and V*Bench (3.15 points).

\begin{table}[H]
\centering
\caption{Selection-policy accuracy and teacher cost. Query ratio is selected/candidate; token ratio is relative to full querying. Bold marks the best 25\%-budget result.}
\label{tab:budget-ablation}
\fontsize{9.5}{10.8}\selectfont
\renewcommand{\arraystretch}{1.12}
\setlength{\tabcolsep}{1.3pt}
\setlength{\aboverulesep}{1.2pt}
\setlength{\belowrulesep}{1.2pt}
\begin{tabular}{
J{\dimexpr0.140\textwidth-2\tabcolsep\relax}
N{\dimexpr0.075\textwidth-2\tabcolsep\relax}
N{\dimexpr0.075\textwidth-2\tabcolsep\relax}
N{\dimexpr0.095\textwidth-2\tabcolsep\relax}
N{\dimexpr0.080\textwidth-2\tabcolsep\relax}
N{\dimexpr0.110\textwidth-2\tabcolsep\relax}
N{\dimexpr0.081\textwidth-2\tabcolsep\relax}
N{\dimexpr0.107\textwidth-2\tabcolsep\relax}
N{\dimexpr0.092\textwidth-2\tabcolsep\relax}
N{\dimexpr0.074\textwidth-2\tabcolsep\relax}
N{\dimexpr0.071\textwidth-2\tabcolsep\relax}
}
\toprule
& \multicolumn{2}{c}{\small\textbf{Teacher Cost}} & \multicolumn{3}{c}{\small\textbf{Fine-Grained Visual Tasks}} & \multicolumn{4}{c}{\small\textbf{Holdout Tasks}} & \\
\cmidrule(lr){2-3}\cmidrule(lr){4-6}\cmidrule(lr){7-10}
\small\textbf{Policy} & \small\textbf{\shortstack{Query\\Ratio}} & \small\textbf{\shortstack{Token\\Ratio}} & \small\textbf{V*Bench} & \small\textbf{\shortstack{Zoom\\Bench}} & \small\textbf{\shortstack{HR-Bench\\4K}} & \small\textbf{MMVP} & \small\textbf{CV-Bench} & \small\textbf{MMStar} & \small\textbf{POPE} & \small\textbf{Avg.} \\
\midrule
Full querying & 100.00 & 100.00 & 87.43 & 58.46 & 82.25 & 78.33 & 85.56 & 77.73 & 88.58 & 79.76 \\
\midrule
Random-25 & 25.00 & 26.28 & 86.43 & 56.51 & 82.88 & 78.33 & 85.77 & 77.00 & 87.83 & 79.25 \\
Entropy-25 & 25.00 & 27.65 & 88.48 & 58.58 & 83.62 & 77.33 & 86.41 & 77.87 & \textbf{89.40} & 80.24 \\
\rowcolor{basrowblue}
\textbf{Learned-25} & 25.00 & \textbf{15.78} & \textbf{90.58} & \textbf{59.88} & \textbf{85.62} & \textbf{79.33} & \textbf{87.63} & \textbf{79.27} & 88.91 & \textbf{81.60} \\
\bottomrule
\end{tabular}
\end{table}

\begin{figure}[H]
    \vspace{-12pt}
    \centering
    \includegraphics[width=\textwidth]{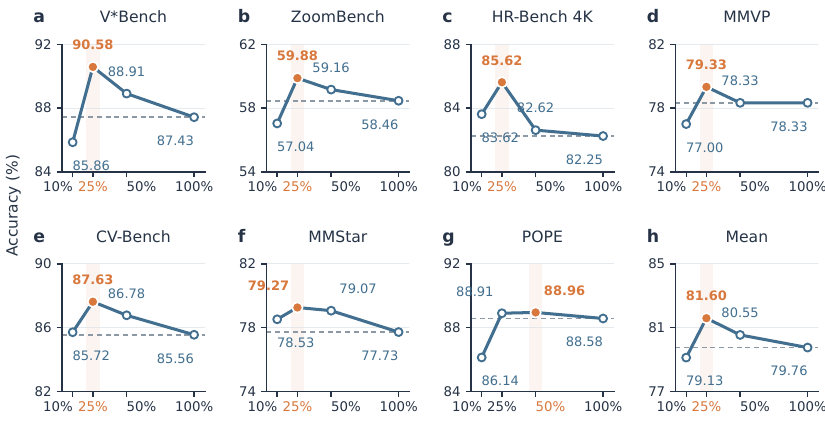}
    \caption{Qwen3.5-4B accuracy across teacher-query budgets on seven benchmarks; panel (h) is their mean. Orange marks each panel's maximum; dashed lines mark full querying.}
    \label{fig:learned-budget}
    \vspace{-10pt}
\end{figure}

Learned-25 exceeded full querying and Random-25 on all seven benchmarks and outperformed Entropy-25 on six, with gains of 1.22--2.10 points; Entropy-25 led on POPE by 0.49 points. Unlike entropy, the learned target also scores confident, support-consistent teacher corrections. At a 25\% query budget, Learned-25 achieved the highest mean accuracy while using 15.78\% of full-query scored tokens (Table~\ref{tab:budget-ablation}). Appendix~\ref{app:teacher-cost} reports detailed token and wall-clock costs.

\noindent\begin{minipage}{\textwidth}
\textbf{Effect of query budget.}
Figure~\ref{fig:learned-budget} compares four budgets on seven benchmarks. The 25\% budget leads six sweeps, while 50\% leads on POPE by 0.05 points. At 10\%, 25\%, 50\%, and 100\%, the seven-benchmark means are 79.13\%, 81.60\%, 80.55\%, and 79.76\%, confirming non-monotonic returns.
\end{minipage}

\begin{figure}[!t]
    \centering
    \includegraphics[width=0.86\textwidth]{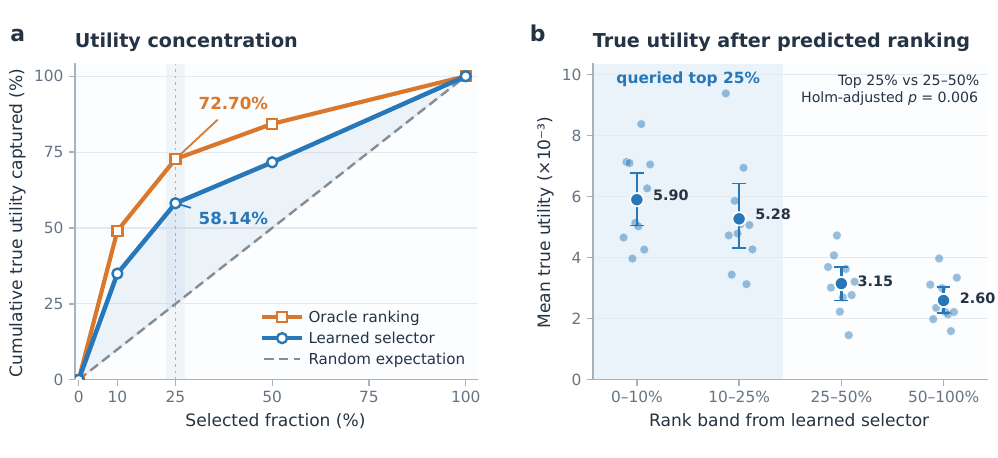}
    \caption{Utility diagnostics over 2,240 candidates from ten batches. (a) Cumulative utility captured by oracle, learned, and random rankings. (b) Mean utility by predicted rank band with 95\% bootstrap confidence intervals; $p$ is Holm-adjusted over three one-sided paired sign-flip tests.}
    \label{fig:utility-quantile}
\end{figure}

\noindent\textbf{Utility concentration and ranking behavior.}
Figure~\ref{fig:utility-quantile} audits 2,240 rollouts across ten batches. At 25\%, learned ranking captures 58.14\% of utility, versus 72.70\% for the oracle and 25\% for random ranking (Figure~\ref{fig:utility-quantile}a). Utility decreases across rank bands; the top-quartile mean is 1.75 times the next quartile ($5.52\times10^{-3}$ versus $3.15\times10^{-3}$; Holm-adjusted $p=0.006$; Figure~\ref{fig:utility-quantile}b). Using training-distribution batches and a frozen final student instead of the unserialized EMA teacher, this audit diagnoses policy behavior rather than out-of-distribution generalization.

\begin{figure}[!t]
    \centering
    \includegraphics[width=0.82\textwidth]{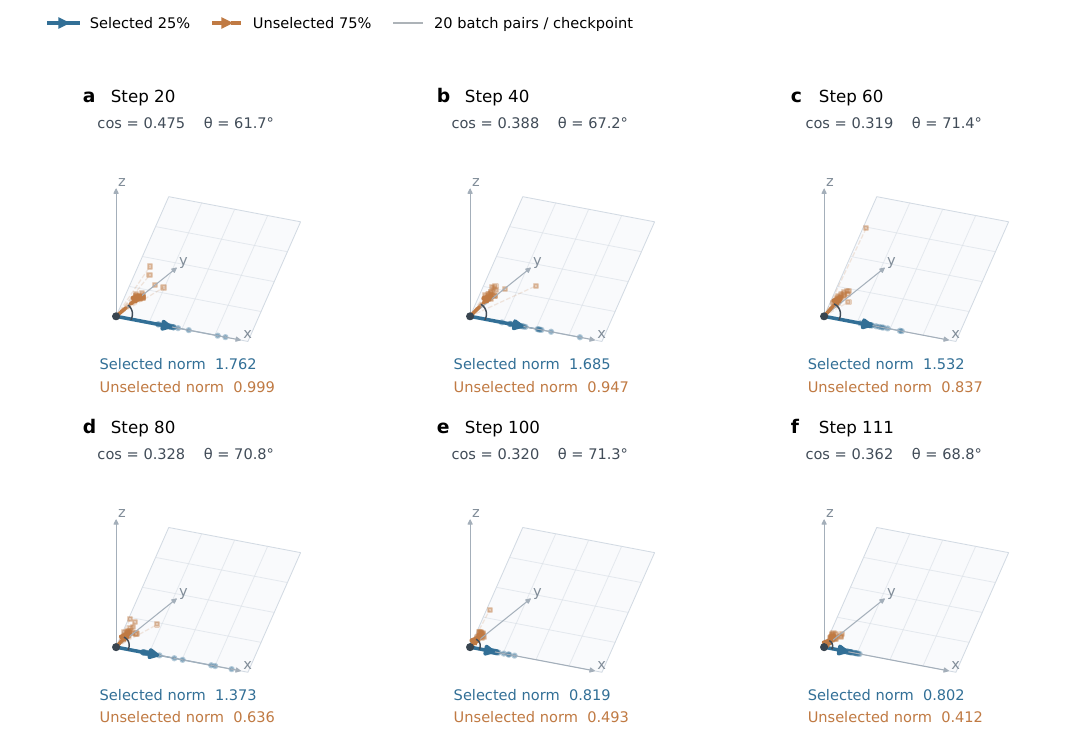}
    \caption{Gradient geometry for Learned-25 at six checkpoints. Arrows show selected and unselected mean gradients; faint vectors show 20 batch pairs per checkpoint. Coordinates are aligned for display.}
    \label{fig:gradient-geometry}
    \vspace{-5pt}
\end{figure}

\noindent\textbf{Gradient geometry.}
Figure~\ref{fig:gradient-geometry} shows that, across six checkpoints, the selected 25\% produced 1.66--2.16$\times$ larger mean OPD gradient norms than the unselected 75\%. Cosine similarities of 0.319--0.475 (angles of $61.7^{\circ}$--$71.4^{\circ}$) indicate partial alignment. These single-trajectory results use 20 batch pairs per checkpoint and do not establish that unselected rollouts are harmful or that 25\% is optimal.
\FloatBarrier

\noindent\textbf{Utility-target and selector-design ablations.}
Table~\ref{tab:selector-components} isolates top-$k$ overlap, teacher confidence, ranking supervision, and random exploration. The full selector achieved the highest seven-benchmark average of 81.60\%. Removing individual components lowered average accuracy by 2.79--3.66 points, with teacher confidence causing the largest drop, consistent with its role in downweighting diffuse teacher distributions. On MMStar, the ablations caused larger drops of 7.80--9.87 points, showing that the gains extended to holdout evaluation, although their magnitude varied by benchmark.

\selectorcomponentstable
\FloatBarrier

\section{Related Work}

\subsection{On-Policy Distillation}

Conventional distillation supervises a student on a fixed data distribution, whereas on-policy distillation evaluates the teacher along student-generated trajectories to reduce the train--inference mismatch. GKD trains on student-generated sequences, while MiniLLM and DistiLLM combine student rollouts with alternative divergence objectives or adaptive on- and off-policy mixtures \citep{agarwal2024gkd,gu2024minillm,ko2024distillm}. Recent self-distillation methods condition the same model on privileged reasoning traces or demonstrations, extending on-policy learning to mathematical reasoning and continual learning \citep{zhao2026selfdistilled,shenfeld2026selfdistillation}. A complementary analysis identifies teacher--student compatibility, novel teacher capability, and dense token-level supervision cost as central factors in successful on-policy distillation \citep{li2026rethinking}. Privileged regional supervision has also been explored for multimodal perception \citep{yuan2026visionopd}. BAS-OPD instead addresses query allocation: under a fixed budget, it selects which rollouts should receive supervision before the teacher forward pass.

\subsection{Fine-Grained Visual Understanding for MLLMs}

Fine-grained MLLM perception is limited when small but decisive evidence is lost during image encoding. High-resolution architectures retain more detail through dynamic tiling, variable visual-token counts, or adaptive image slicing, as in InternVL~1.5, Qwen2-VL, and LLaVA-UHD \citep{chen2024internvl,wang2024qwen2vl,guo2024llavauhd}. Inference-time methods such as V* and DC$^2$ instead search, partition, or retrieve relevant regions for each query \citep{wu2024vstar,wang2025dc2}. These approaches improve access to local evidence, but increase visual processing or introduce additional inference steps. BAS-OPD is complementary: it changes neither the visual architecture nor the inference procedure, but allocates privileged regional supervision selectively during training and retains single-pass full-image inference.

\section{Conclusion and Limitations}

We introduced BAS-OPD, which allocates crop-teacher supervision under a fixed query budget using online utility estimates. Learned-25 raised the Qwen3.5-4B seven-benchmark average from 79.76\% to 81.60\% while querying 25\% of candidates and using 15.78\% of full-query scored tokens; BAS-OPD 9B reached 84.23\%. BAS-OPD thus improves the training accuracy--cost trade-off while preserving single-pass full-image inference.

\noindent\textbf{Limitations.}

Selection analyses use Qwen3.5-4B and one seed, and the budget sweep covers seven benchmarks; broader seeds, backbones, and matched 9B ablations remain untested. Wall-clock results depend on implementation and hardware utilization. Because the EMA teacher was not serialized, the frozen-selector audit measures training-distribution subset retrieval rather than out-of-distribution generalization.

\label{sec:main-end}

\clearpage
\bibliography{iclr2027_conference}
\bibliographystyle{iclr2027_conference}

\clearpage
\appendix
\section{Implementation Details}

\subsection{Random and entropy selection baselines}
\label{app:selection-baselines}

Random and student-entropy selection use the same query-budget interface as
learned selection, but neither trains a utility predictor.

\textbf{Random-25} samples $K=\lfloor0.25|\mathcal{C}|\rfloor$ eligible rollouts
uniformly without replacement at each training step. Sampling uses a CPU
random permutation with a seed derived from the selector seed and training
step. Only these selected rollouts are sent to the crop teacher. The policy
has no trainable selector and requires no additional student forward pass.

\textbf{Entropy-25} is a fixed student-uncertainty heuristic. When token entropy is
already available from the rollout path, each candidate is scored by its mean
valid-token entropy. Otherwise, the selector uses mean sampled-token negative
log-likelihood (NLL) from the stored rollout log-probabilities. The top-$K$
rows are queried. This policy adds no trainable parameters and requires no
additional student forward pass.

\subsection{Predictor configuration and feature handling}
\label{app:bas-features}

Algorithm~\ref{alg:bas-opd} (line~\ref{alg:features}) forms the detached feature
vector in Eq.~\ref{eq:bas-features}. The sampled-token negative log-likelihood
is $\ell_{i,t}=-\log p_{\mathrm{old}}(y_{i,t}\mid x_i,q_i,y_{i,<t})$;
$\overline{\ell}_i$ and $Q_{0.9}(\ell_i)$ are its response-level mean and
90th percentile. The remaining features are mean token entropy
$\overline{H}_i$ when available, response length $T_i$, prompt length $L_i$,
and crop-to-full-image area ratio $a_i$.

The predictor $f_{\psi}$ consists of layer normalization, one hidden linear
layer with a SiLU activation, and a scalar output. The operator
$\operatorname{Normalize}(\cdot)$ uses running feature statistics, updated
on crop-eligible candidates during feature extraction and held fixed during
replay updates.

Table~\ref{tab:supp-selector-config} gives the selector settings shared by
Learned-10, Learned-25, and Learned-50.
The predictor runs on the CPU and ranks candidates by predicted utility,
without dividing by response-token cost. Running feature normalization uses
exponential mean and variance with numerical constant $10^{-6}$.
Unavailable entropy, prompt-length, or crop-area values are represented by
zero. If separate teacher top-$k$ indices are unavailable, the implementation
sets the overlap factor to one.

\begin{table}[!htbp]
\centering
\caption{Learned-selector configuration shared across query budgets.}
\label{tab:supp-selector-config}
\fontsize{9.5}{10.8}\selectfont
\renewcommand{\arraystretch}{1.12}
\setlength{\tabcolsep}{4pt}
\setlength{\aboverulesep}{1.2pt}
\setlength{\belowrulesep}{1.2pt}
\begin{tabular}{
J{\dimexpr0.620\textwidth-2\tabcolsep\relax}
J{\dimexpr0.380\textwidth-2\tabcolsep\relax}
}
\toprule
\rowcolor{basgroupgray}
\textbf{Hyperparameter} & \textbf{Value} \\
\midrule
Feature dimension / hidden dimension & 6 / 32 \\
Selector optimizer / learning rate & Adam / $1\times10^{-3}$ \\
Running-normalizer momentum & 0.99 \\
\midrule
Replay capacity / start threshold & 512 / 96 labels \\
Replay minibatch size & 256 \\
Selector updates per training step & 2 \\
Replay age half-life & 10 steps \\
\midrule
Pairwise ranking weight & 0.2 \\
Minimum utility gap for a pair & $1\times10^{-6}$ \\
\midrule
Exploration ratio & 0.10 \\
Selector seed & 0 \\
Score mode & predicted utility \\
\bottomrule
\end{tabular}
\end{table}

\subsection{Selector training and update handling}
\label{app:selector-training}

Algorithm~\ref{alg:bas-opd} (line~\ref{alg:selector-update}) stores observed
feature--utility pairs in a bounded first-in, first-out replay buffer together
with their collection steps. Replay updates require newly observed valid pairs
and the minimum buffer size specified in Table~\ref{tab:supp-selector-config}.
Each update samples a minibatch uniformly without replacement, capped by the
current buffer size. At training step $s$, an item collected at $s_i$ receives
the recency weight $w_i=2^{-(s-s_i)/h}$, where $h$ is the age half-life.

For each replay update, we normalize stored features using the current
statistics without changing those statistics, and recompute $\hat{u}_i$
with the current predictor while retaining gradients to $\psi$. The weighted
Huber term in Eq.~\ref{eq:bas-selector-loss} is
\begin{equation}
\mathcal{L}_{\mathrm{reg}}
=\frac{\sum_i w_i\,\operatorname{Huber}(\hat{u}_i-u_i)}{\sum_i w_i}.
\label{eq:supp-selector-regression}
\end{equation}
The ranking term $\mathcal{L}_{\mathrm{rank}}$ trains the ordering of labeled
samples. Each pair $(i,j)$ with $|u_i-u_j|>10^{-6}$ contributes
$\operatorname{softplus}[-\operatorname{sgn}(u_i-u_j)(\hat{u}_i-\hat{u}_j)]$,
weighted by $\sqrt{w_iw_j}$. Features and targets are detached, and the
selector has a separate optimizer, so $\mathcal{L}_{\mathrm{sel}}$ does not
update the MLLM.

The denominator of Eq.~\ref{eq:bas-distillation} is clamped to at least one.
When $\mathcal{S}=\varnothing$, crop-teacher scoring is skipped and the
distillation term is zero; the remaining updates follow the standard OPD
optimizer and teacher-update rules. The teacher EMA follows a valid student
update (Algorithm~\ref{alg:bas-opd}, line~\ref{alg:distill}). An empty selected
set provides no new utility labels and therefore does not trigger a replay
update.

\section{Training and Evaluation Details}

\subsection{Training configuration}

Table~\ref{tab:supp-training-config} records the batch configuration and step
counts for the four retained policy runs. The maximum model length is 9,216
tokens; the optimizer and remaining training settings are specified in the
main experimental setup.

\begin{table}[!htbp]
\centering
\caption{Training configurations for the four selection-policy runs.}
\label{tab:supp-training-config}
\fontsize{9.5}{10.8}\selectfont
\renewcommand{\arraystretch}{1.12}
\setlength{\tabcolsep}{4pt}
\setlength{\aboverulesep}{1.2pt}
\setlength{\belowrulesep}{1.2pt}
\begin{tabular}{
J{\dimexpr0.320\textwidth-2\tabcolsep\relax}
N{\dimexpr0.170\textwidth-2\tabcolsep\relax}
N{\dimexpr0.170\textwidth-2\tabcolsep\relax}
N{\dimexpr0.170\textwidth-2\tabcolsep\relax}
N{\dimexpr0.170\textwidth-2\tabcolsep\relax}
}
\toprule
\rowcolor{basgroupgray}
\textbf{Configuration} & \textbf{Full querying} & \textbf{Random-25} & \textbf{Entropy-25} & \textbf{Learned-25} \\
\midrule
Prompt batch & 28 & 28 & 28 & 28 \\
Rollouts per prompt & 8 & 8 & 8 & 8 \\
Candidates per step & 224 & 224 & 224 & 224 \\
Selected per step & 224 & 56 & 56 & 56 \\
Completed steps & 111 & 111 & 111 & 111 \\
Internal query ratio & 100\% & 25\% & 25\% & 25\% \\
Training seed & 42 & 42 & 42 & 42 \\
\bottomrule
\end{tabular}
\end{table}

\subsection{Evaluation protocol}

Generation uses temperature 0 and seed 42. Benchmark-native option or
exact-answer parsing is used when available; remaining free-form answers are
judged with GPT-5.5 at temperature 0. CV-Bench is reported as the macro-average
of its 2D and 3D accuracy categories, and POPE is reported by accuracy in the
policy comparison. The POPE evaluation contains 9,000 samples.

\section{Teacher-Supervision Cost}
\label{app:teacher-cost}

Training cost is recorded directly as the number of selected teacher calls,
$\lvert\mathcal{S}\rvert$, and the number of response tokens scored by the
teacher, $\sum_{i\in\mathcal{S}}T_i$.

\begin{figure}[!htbp]
    \centering
    \includegraphics[width=\textwidth]{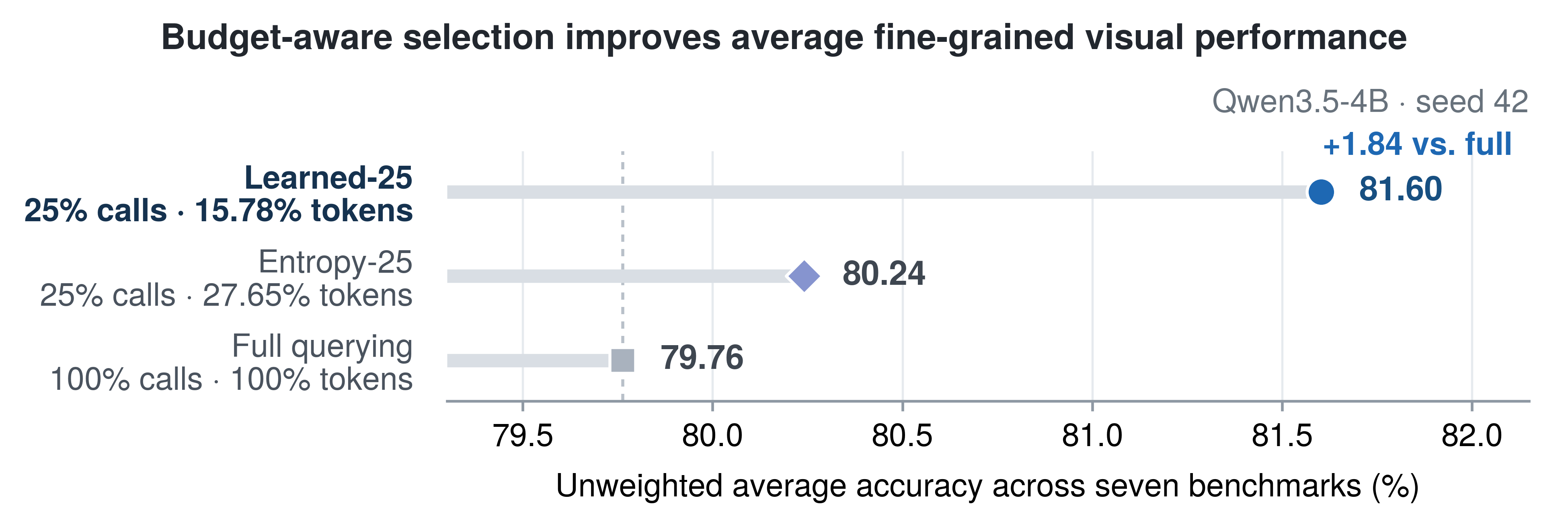}
    \caption{Accuracy--cost comparison on Qwen3.5-4B (seed 42). Points show the unweighted mean accuracy across seven benchmarks. Row labels report the teacher-query budget and the ratio of scored tokens to full querying.}
    \label{fig:accuracy-cost}
\end{figure}

Figure~\ref{fig:accuracy-cost} summarizes the average accuracy and teacher cost.
Tables~\ref{tab:supp-teacher-cost} and~\ref{tab:supp-wallclock} provide the
recorded supervision counts and observed training times underlying the main
accuracy--cost comparison.

\noindent\textbf{Teacher-scored tokens.}
Random-25 and Entropy-25 used 26.28\% and 27.65\% of full-query teacher-scored tokens, whereas Learned-25 used 15.78\% (Table~\ref{tab:budget-ablation}). These reductions of 10.50 and 11.87 percentage points reflect shorter selected responses, since all three policies queried one quarter of their candidates. Learned-25 therefore combined the highest aggregate accuracy with the fewest scored tokens.

\noindent\textbf{Observed training time.}
Wall-clock savings were smaller: Learned-25 reduced accumulated teacher-forward time by 16.02\% and total step time by 9.84\% relative to full querying. Student generation, communication, data loading, and rollout scheduling remain substantial costs. We therefore report teacher calls, scored tokens, teacher-forward time, and total step time together, with measurement details and comparability limits in Table~\ref{tab:supp-wallclock}.

\begin{table}[!htbp]
\centering
\caption{Measured teacher-supervision cost. Token ratios use Full querying as
the denominator. For Learned-25, the 25\% budget is computed directly from
the recorded aggregate counts as $6{,}216/24{,}864$.}
\label{tab:supp-teacher-cost}
\fontsize{9.5}{10.8}\selectfont
\renewcommand{\arraystretch}{1.12}
\setlength{\tabcolsep}{3pt}
\setlength{\aboverulesep}{1.2pt}
\setlength{\belowrulesep}{1.2pt}
\begin{tabular}{
J{\dimexpr0.180\textwidth-2\tabcolsep\relax}
N{\dimexpr0.140\textwidth-2\tabcolsep\relax}
N{\dimexpr0.130\textwidth-2\tabcolsep\relax}
N{\dimexpr0.140\textwidth-2\tabcolsep\relax}
N{\dimexpr0.165\textwidth-2\tabcolsep\relax}
N{\dimexpr0.145\textwidth-2\tabcolsep\relax}
N{\dimexpr0.100\textwidth-2\tabcolsep\relax}
}
\toprule
\rowcolor{basgroupgray}
\textbf{Policy} & \textbf{Candidates} & \textbf{\shortstack{Teacher\\calls}} & \textbf{\shortstack{Query\\ratio}} & \textbf{\shortstack{Scored\\tokens}} & \textbf{\shortstack{Tokens vs.\\Full}} & \textbf{\shortstack{Max\\overrun}} \\
\midrule
Full querying & 24,864 & 24,864 & 100.00\% & 2,167,850 & 100.00\% & 0 \\
Random-25 & 24,864 & 6,216 & 25.00\% & 569,674 & 26.28\% & 0 \\
Entropy-25 & 24,864 & 6,216 & 25.00\% & 599,321 & 27.65\% & 0 \\
\rowcolor{basrowblue}
\textbf{Learned-25} & 24,864 & 6,216 & 25.00\% & 342,089 & 15.78\% & 0 \\
\bottomrule
\end{tabular}
\end{table}

\begin{table}[!htbp]
\centering
\caption{Observed training time on the original machines. These measurements
are secondary because code path, device utilization, and rollout scheduling
are not fully controlled across runs.}
\label{tab:supp-wallclock}
\fontsize{9.5}{10.8}\selectfont
\renewcommand{\arraystretch}{1.12}
\setlength{\tabcolsep}{3pt}
\setlength{\aboverulesep}{1.2pt}
\setlength{\belowrulesep}{1.2pt}
\begin{tabular}{
J{\dimexpr0.200\textwidth-2\tabcolsep\relax}
N{\dimexpr0.220\textwidth-2\tabcolsep\relax}
N{\dimexpr0.190\textwidth-2\tabcolsep\relax}
N{\dimexpr0.180\textwidth-2\tabcolsep\relax}
N{\dimexpr0.210\textwidth-2\tabcolsep\relax}
}
\toprule
\rowcolor{basgroupgray}
\textbf{Policy} & \textbf{\shortstack{Teacher\\forward sum}} & \textbf{\shortstack{Mean\\teacher/step}} & \textbf{Mean step} & \textbf{\shortstack{Total\\step time}} \\
\midrule
Full querying & 5,174.38\,s & 46.62\,s & 499.88\,s & 55,487.15\,s \\
Random-25 & 4,705.55\,s & 42.39\,s & 468.95\,s & 52,053.55\,s \\
Entropy-25 & 4,690.39\,s & 42.26\,s & 469.30\,s & 52,092.85\,s \\
\rowcolor{basrowblue}
\textbf{Learned-25} & 4,345.38\,s & 39.15\,s & 450.70\,s & 50,027.87\,s \\
\bottomrule
\end{tabular}
\end{table}

\end{document}